\documentclass[a4paper]{svmult}

\usepackage{type1cm}        
\usepackage{makeidx}         
\usepackage{graphicx}        
\usepackage{multicol}        
\usepackage[bottom]{footmisc}

\usepackage{newtxtext}       %
\usepackage{newtxmath}       

\usepackage{mwe}

\makeindex             

\newcommand{\sensorobs}{\mathcal{Z}^i_t}
\newcommand{\sensorobsi}{\mathcal{Z}^i_t}

\newcommand{\gmmparams}{\Theta}
\newcommand{\mean}{\boldsymbol{\mu}}
\newcommand{\cov}{\boldsymbol{\Sigma}}
\newcommand{\weight}{\pi}
\newcommand{\gmm}{\gmmparams_{\sensorobs}}
\newcommand{\keyframegmm}{\hat{\gmmparams}_{\sensorobs}}
\newcommand{\keyframeroboti}{\hat{\gmmparams}_{\sensorobsi}}
\newcommand{\poserobotn}{\mathbf{S}^i_t}
\newcommand{\poseroboti}{\mathbf{S}^i_t}

\newcommand{\rot}{\mathbf{R}}
\newcommand{\trans}{\mathbf{x}}
\newcommand{\point}{\mathbf{p}}
\newcommand{\ogmapvar}{m}
\newcommand{\localog}{\mathbf{\ogmapvar}^i_t}
\newcommand{\localogj}{\mathbf{\ogmapvar}^j_t}

\input{preamble}

\begin{document}
\mainmatter

\title*{Rapid and High-Fidelity Subsurface Exploration with Multiple Aerial Robots}
\titlerunning{Rapid and High-Fidelity Subsurface Exploration with Multiple Aerial Robots}  
\author{Kshitij Goel \and Wennie Tabib \and Nathan Michael}
\authorrunning{Goel et al.} 

\institute{The authors are with the Robotics Institute, Carnegie Mellon University, Pittsburgh, PA 15213 USA (e-mail: \{kgoel1,wtabib,nmichael\}@andrew.cmu.edu)}

%
%
\maketitle

\abstract{This paper develops a communication-efficient distributed
mapping approach for rapid exploration of a cave by a multi-robot
team. Subsurface planetary exploration is an unsolved problem
challenged by communication, power, and compute constraints.
Prior works have addressed the problems of rapid exploration
and leveraging multiple systems to increase exploration rate;
however, communication considerations have been left largely
unaddressed. This paper bridges this gap in the state of the
art by developing distributed perceptual modeling that enables
high-fidelity mapping while remaining amenable to low-bandwidth
communication channels. The approach yields significant
gains in exploration rate for multi-robot teams as
compared to state-of-the-art approaches. The work is
evaluated through simulation studies and hardware experiments
in a wild cave in West Virginia.}

\section{Introduction}\label{sec:intro}
Planetary exploration has benefited from advancements in robotics through
automation of data collection for planetary science and robotic precursor
missions for human space exploration~\citep{board2012vision}. To date, robotic
precursor missions have engaged in surface exploration of
Mars~\citep{maimone2007two} but have not explored subsurface environments
despite the potential geological and astrobiological significance of these
domains~\citep{phillips2020macie,stamenkovic2019next}. As a result, robotic
subsurface exploration has been identified as a key technology for future
missions to these planets~\citep{miranda20202020}. Autonomous navigation and
high-resolution perceptual modeling are critical needs in the context of
subsurface planetary exploration~\citep{whittaker2014exploration}. A
challenge of operating in subsurface environments is communicating to a surface
station. Communication may be limited or impossible due to the inability of
radio waves to penetrate rock, impeding data relay to Earth, so compact
data transmission is critical. Operating on planets far from
Earth introduces additional restrictions on power and compute that may
be mitigated by leveraging multiple robots to increase coverage in
spatially expansive environments \citep{cortes2017coordinated}.
This work addresses a key challenge
for planetary exploration: enabling rapid multi-robot exploration in subsurface
environments by leveraging a perceptual modeling framework amenable to
low-bandwidth communication while remaining high-fidelity.
\begin{figure}[t]
  \centering
  \includegraphics[width=\textwidth]{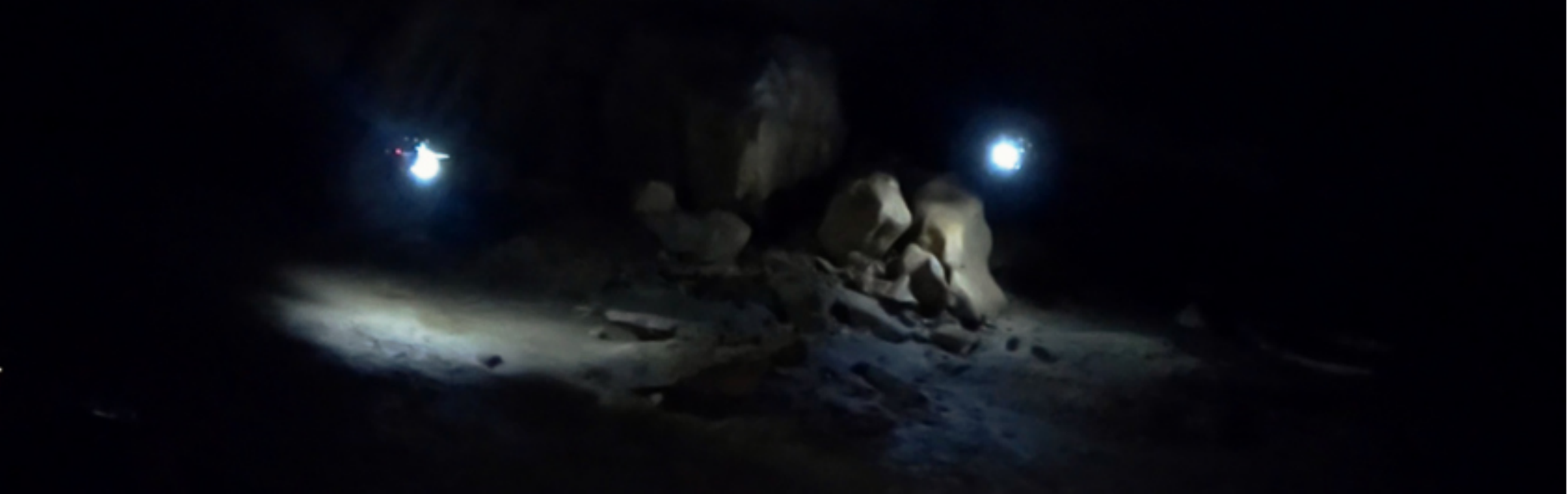}
  \caption{\label{fig:intro}Cave exploration with two aerial robots in West
    Virginia, USA. A video of the flight can be accessed at the following link: \texttt{https://youtu.be/osko8EKKZUM}.}
\end{figure}

Exploration frameworks cannot assume \emph{a priori} knowledge about the
structure of the environment so the exploration system must operate with unknown
locomotion constraints. Aerial robots have recently been leveraged to mitigate
these constraints in the subterranean domain~\citep{Tabib2020TRO} and considered
for subsurface mapping on Mars~\citep{phillips2020macie}. In this work, we
consider aerial robots operating in a cave on Earth
(\cref{fig:intro}) as an analog scenario for subsurface exploration on Mars.
These robots are often limited by size, weight, and power (SWaP)
constraints~\citep{Tabib2020TRO}. Energy constraints on these platforms impose
limits on flight endurance necessitating rapid exploration, since the existence
of a replenishment infrastructure in the planetary exploration context cannot be
guaranteed at these sites~\citep{whittaker2014exploration}. Several frameworks
for rapid exploration have been proposed that either use a single fast-moving
aerial
robot~\citep{cieslewski2017,goel2019fsr,Dai2020,dharmadhikari2020motion}
or multiple slow-moving aerial robots~\citep{cesare2015multi,corah2018ral}; however, a real-world deployable framework that combines the elements from both
is desirable. Such a deployment can potentially be realized by sending multiple
aerial robots (``daughtercraft'') from a lander (``mothership'') to perform
rapid, effective, and affordable high-resolution mapping of the target
environment, similar to the concept surface mission studied
by~\citet{matthies2017niac} for Titan. To this end, a distributed perceptual
modeling framework that provides communication-efficient map sharing can
enable the daughtercraft team to improve the rate of exploration while
transmitting scientific data to the mothership and Earth. The experimental
evaluation in this work (\cref{sec:exp_results}) is motivated by this concept of
operations.

\noindent
\textbf{Related Work:} With the ongoing DARPA Subterranean
Challenge~\citep{subtwebsite}, there is an increased interest in deploying a
team of robots in cave networks. Although existing systems have not shown
autonomous operations in a cave, multi-robot exploration systems have been
proposed for mine and tunnel environments. These environments differ from
cave environments because in most cases the terrain is flat. Notably,
teams competing in the challenge have identified that accounting for limited
availability of communication resources within the exploration framework is a
key milestone for future
work~\citep{ebadi2020lamp,dang2020graph,rouvcek2019darpa}. \citet{ebadi2020lamp}
state that the communication bottlenecks faced during mapping with a multi-robot
system due to the use of downsampled point clouds can be addressed by map
compression techniques or compact representations for motion planning.
\citet{dang2020graph} use the state-of-the-art, memory-efficient
OctoMap~\citep{hornung2013octomap} approach for map representation but mention
efficient map sharing as one of the future challenges. \citet{rouvcek2019darpa}
use elevation maps for mapping but only on wheeled and ground robots because the
transmission of these maps requires a physically large communication module. Furthermore,
the speeds of robots are constrained for these systems. These shared challenges
indicate a gap in the state-of-art for communication-efficient distributed
mapping methods in rapid aerial multi-robot exploration systems for subterranean
domains.~\citet{corah2018ral} highlight the benefits of a distributed mapping
strategy that exploits the compactness of Gaussian Mixture Models (GMMs)
relative to the occupancy grid approach~\citep{elfes1989}. However, the approach is
computationally prohibitive for real-world deployment, limits robot
speeds, and the effects of communication constraints on the exploration
performance of the robot team are not discussed.


\noindent
\textbf{Contributions:} We build upon prior single-robot
works~\citep{Tabib2020TRO,goel2019fsr} to develop multi-robot exploration with
the following contributions: (1) a GMM-based distributed
mapping approach and occupancy reconstruction for information-theoretic motion
planning; (2) an evaluation of the fidelity and memory consumption of the
approach against OctoMap~\citep{hornung2013octomap} and occupancy grid
mapping~\citep{elfes1989}; and (3) a simulation study on the effects of
constrained communication on the exploration rate of a two-robot team. All
contributions are presented in the context of caves.


\section{Technical Approach}\label{sec:approach}
\begin{figure}[t]
  \centering
  \includegraphics[width=0.7\linewidth]{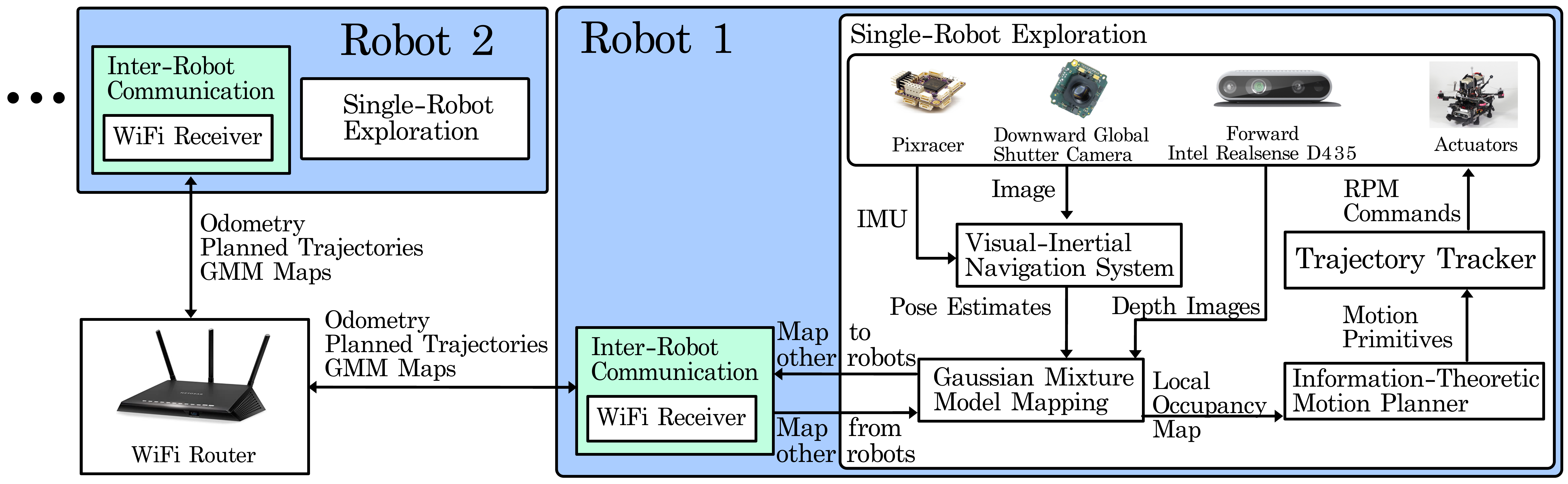}%
  \includegraphics[width=0.3\linewidth]{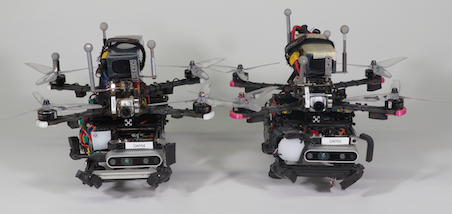}
  \caption{(Left) Overview of the rapid multi-robot exploration framework and
    (Right) aerial systems used in experiments in this work.
    \label{fig:overview}
  }
\end{figure}
An overview of the system is shown in~\cref{fig:overview}. Each robot
is equipped with single-robot exploration and inter-robot
communication modules.  The exploration module consists of four major
subsystems: GMM mapping, information-theoretic motion planning,
visual-inertial state estimation, and trajectory tracking. The
inter-robot communication module enables sharing information between
robots or other computers on the network. The GMM mapping and
planning subsystems together with the communication module constitute
distributed mapping (\cref{ssec:dist_mapping}) and multi-robot
planning (\cref{ssec:planning}), respectively. In this section, the
following mathematical notation is used: lower-case letters represent
scalar values, lower-case bold letters represent vectors, upper-case
bold letters represent matrices, and script letters represent sets.

\subsection{GMM-based Distributed Mapping\label{ssec:dist_mapping}}
\begin{figure}[t]
  \centering
  \subfloat[\label{sfig:data}]{\includegraphics[width=0.33\linewidth,trim=360 100 120 130,clip]{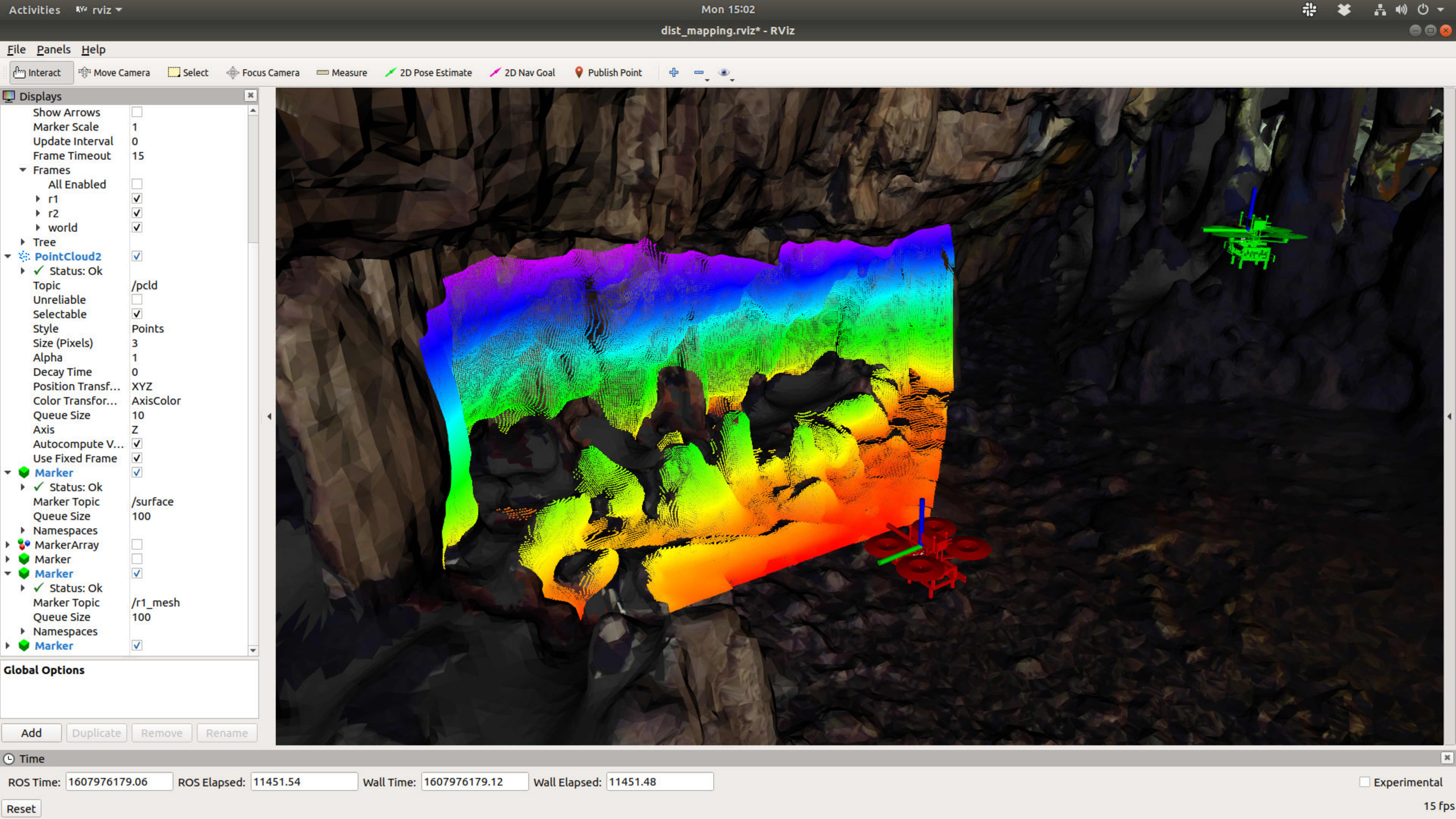}}%
  \subfloat[\label{sfig:gmm_sent}]{\includegraphics[width=0.33\linewidth,trim=360 100 120 130,clip]{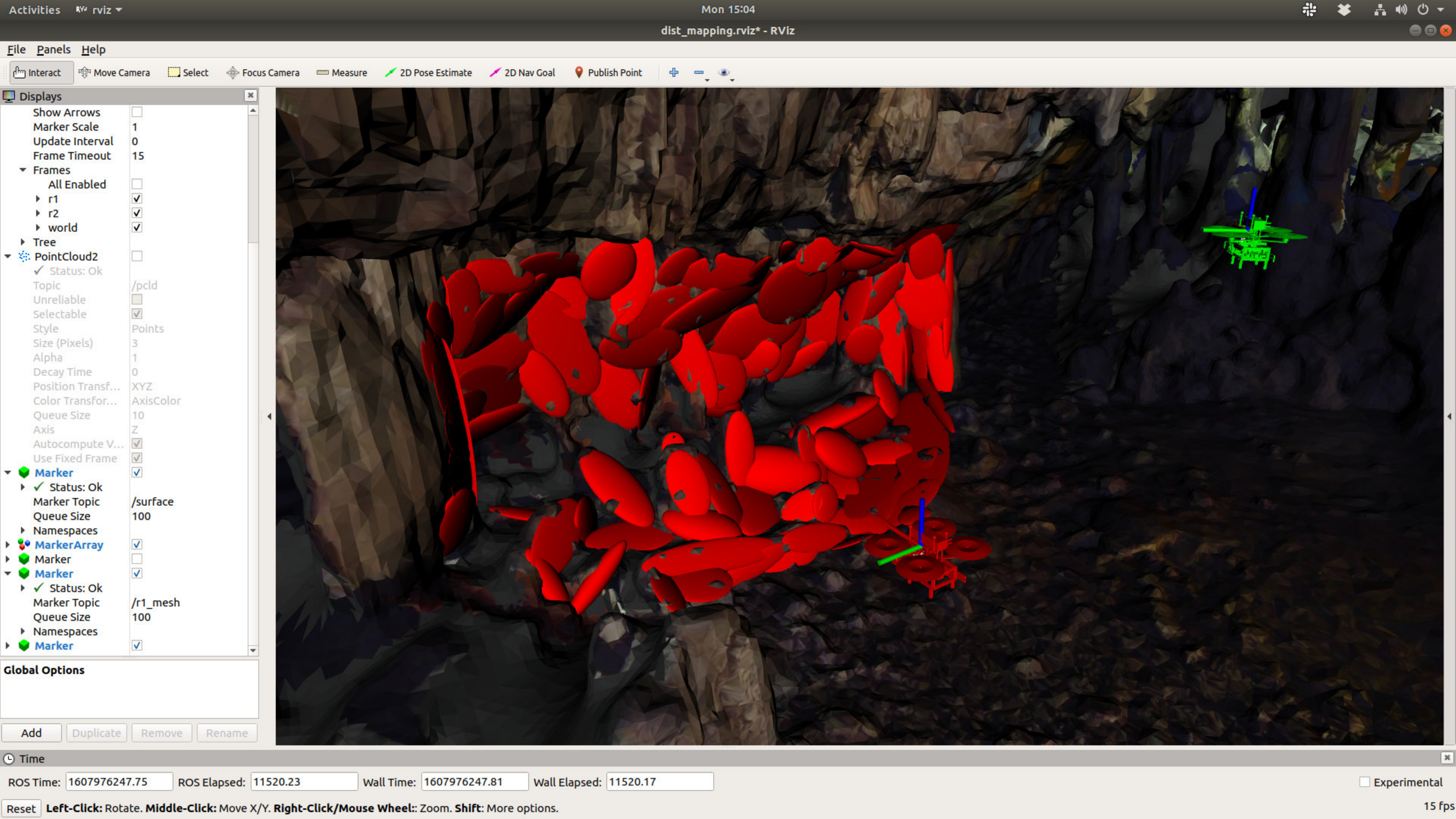}}%
  \subfloat[\label{sfig:map_update}]{\includegraphics[width=0.33\linewidth,trim=360 100 120 130,clip]{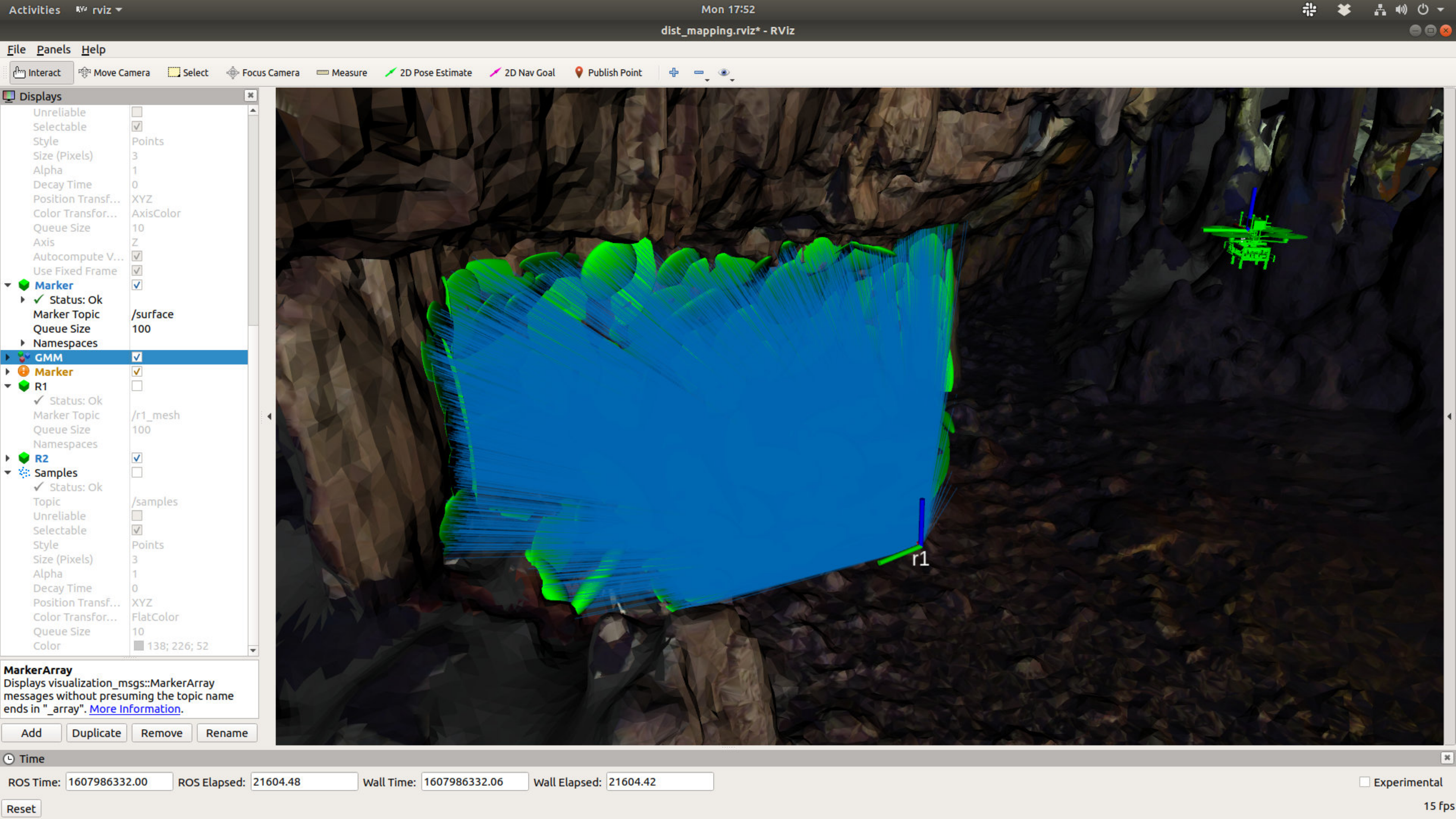}}%
  \caption{\label{fig:distributed_mapping}Overview of the distributed
    mapping approach. \protect\subref{sfig:data} Robot $i$ shown in red, takes
    a sensor observation shown in colors varying from red to purple and
    \protect\subref{sfig:gmm_sent} learns a GMM (shown in red). If the GMM is
    determined to be a keyframe both the GMM and sensor pose are transmitted to 
    robot $j$ (shown in green).
    \protect\subref{sfig:map_update} The GMM and the sensor pose are transformed
    into the frame of robot $j$ and used to update the occupancy.}
\end{figure}

This section details the distributed mapping approach to share environment
models between robots. Consider a team of $N$ robots. At
timestep $t$ robot $i \in N$ receives the depth sensor observation, $\sensorobs$,
which represents a set of points. A Gaussian mixture model (GMM) is learned from
these points following the approach from~\cite{Tabib2020TRO}. The GMM is
parameterized by $\gmmparams = \{ \weight_m, \mean_m, \cov_m \}_{m=1}^M$ where
$\mean_m \in \mathbb{R}^3$ is a mean, $\cov_m \in \mathbb{R}^{3 \times 3}$ is a
covariance, and $\weight_m \in \mathbb{R}$ is a weight such that $\sum_{m=1}^M
\weight_m = 1$. A GMM representing point set $\sensorobs$ is denoted as $\gmm$.

\textbf{Keyframe GMMs:} To reduce redundant observations, keyframe GMMs
are identified for transmission to other robots.  A keyframe GMM,
$\keyframegmm$, is determined by approximating the field of view for
the current sensor observation as a rectangular pyramid and
calculating the overlapping volume with other keyframe fields of view.
If the volume is smaller than a user-defined threshold, $\lambda$, the
sensor observation is considered to be a keyframe. $\keyframegmm$ and
the sensor pose, $\poserobotn \in$ SE(3), are transmitted to the other
robots or computers on the network.

Each robot maintains its own environment representation and relative initial transforms
between robots are assumed to be known.  When robot $j$ receives
$\keyframeroboti$, it is received in the frame of robot $i$.
To transform it into the frame of robot $j$, the relative initial
rotation $\rot^{ji}_0 \in \mathbb{R}^{3
\times 3}$ and translation $\trans^{ji}_0 \in \mathbb{R}^{3}$ parameters
are applied to the means and covariances of the distribution using
the following equations.
\begin{align}
  \mean^j = \rot^{ji}_0 \mean^i + \trans^{ji}_0 ~~~~~~~~~~~~ \cov^j = \rot^{ji}_0 \cov^i (\rot^{ji}_0)^T,
\end{align}
The transformed GMM is incorporated into robot $j$'s existing GMM map
following the approach from~\cite{tabibRSS2019,Tabib2020TRO}.

\textbf{Occupancy Reconstruction:} A local occupancy grid map
$\localog$ is maintained and centered around the robot's current
position $\trans^i_{t}$ for use in information-theoretic motion
planning. To generate $\localog$, a number of points $\point \in
\mathbb{R}^{3}$ equal to the support size, or number of points
used to learn the distribution, is sampled and raytraced to the sensor pose
$\trans^i_{t}$. The probability of occupancy along the ray is
updated.

\textbf{Multi-robot Map Updates:} Care must be taken to update
$\localogj$ when receiving $\keyframeroboti$. In addition to applying the
transformation parameters so that $\keyframeroboti$ is transformed into the
frame of robot $j$, $\localogj$ must also be updated by sampling points from the
transformed $\keyframeroboti$ and raytracing through $\localogj$ to the sensor
pose, $\poseroboti$, which must also be transformed into the frame of robot
$j$. This ensures the occupancy is updated with observations from both robots. A
visualization of this is shown in~\cref{fig:distributed_mapping}. Robot $i$
takes a sensor observation (\cref{sfig:data}) and learns $\keyframeroboti$ (\cref{sfig:gmm_sent}). This
keyframe GMM is transmitted to robot $j$, transformed into the frame of robot
$j$, and then used to update $\localogj$ (\cref{sfig:map_update}).


\subsection{Planning for Rapid Multi-Robot Exploration\label{ssec:planning}}
Robot $i$ uses $\localog$ for information-theoretic receding-horizon planning
via the strategy presented in~\citep{goel2019fsr}, which accounts for perception latencies and kinodynamic
constraints of the robot. The approach uses Monte Carlo tree search
(MCTS)~\citep{chaslot2010} to evaluate the Cauchy-Schwarz Quadratic Mutual
Information (CSQMI)~\citep{charrow2015icra} for a set of motion primitives over
a user-specified time horizon. An informative primitive sequence is
selected that maximizes the CSQMI over the MCTS tree.
Safety is ensured by checking for collisions with the environment.

The informative trajectories are shared with other robots and
inter-robot collision avoidance is enabled through a standard
priority-based collision checker assuming a cylindrical robot
model~\citep{cai2007collision}. The priorities are assigned manually before the
exploration run and remain constant throughout. To reduce the computational
complexity for lower priority robots, three optimizations are applied. First,
the collision checking is only active when a pair of robots are within a
pre-specified radius. To enable this on each robot without assuming a
centralized oracle, the robots share odometry information at a sufficiently high
rate ($\SI{10}{\hertz}$) compared to the planning frequency ($\SI{1}{\hertz}$).
Second, the number of cylinders sampled over the planned trajectory is limited
to a pre-specified maximum to cap the number of cylinder-cylinder collision
checks. This maximum value and the associated cylinder collision radius
are selected conservatively based on the length of the motion primitive
assuming the robot starts at hover and achieves a top speed at the
endpoint. Third, for each robot the collision checks are performed only with
the candidate motion primitive and the associated stopping motion primitive at the first depth of
the MCTS tree because each depth of the tree is of a sufficiently long duration
($\SI{2}{\second}$) as compared to the planning time ($\SI{1}{\second}$).
The inter-robot collision checker is used in the constrained-bandwidth simulation study
(\cref{sssec:constrained_bandwidth}).

\section{Experimental Design and Results}\label{sec:exp_results}
The experimental evaluation is motivated through a concept of operations for a
multi-robot exploration mission in a Martian cave. Two robotic systems explore a
Martian cave, transmit their maps to a surface station, which serves
as a relay to an orbiter, and the orbiter transmits the data to operators on
Earth. Three evaluations are conducted to quantify the system performance
through this concept of operations: first, the perceptual fidelity and memory
usage of the map is compared to state-of-the-art approaches in a representative
cave environment (\cref{sssec:perceptual_detail}); second, a hardware experiment
is demonstrated with two rapidly exploring aerial systems and the communication
requirement for each mapping approach is compared (\cref{sssec:exploration});
and third, a simulation study is conducted to study the effects of the bandwidth
constraints on exploration performance (\cref{sssec:constrained_bandwidth}).

To correctly analyze the performance of the simulation study, the
bottleneck in data transmission rate is identified and bounds on the
rates are determined.
In this scenario, data is transferred between the subterranean robot
and surface station\footnote{\citet{whittaker2014exploration} suggest
the use of either very low frequency (VLF) radios or magneto-inductive
(MI) links to achieve limited data rate through thick layers of
rock. The MI links in particular can provide approximately
$20$-$\SI{25}{\meter}$ dry soil penetration at channel capacity
ranging from $0.1$-$\SI{0.25}{\mega\bit\per\second}$ when using small
antennas (coils)~\citep{kisseleff2018survey}. In the results presented
in~\cref{sssec:constrained_bandwidth}, it is assumed that the robots
could be equipped with these MI links.},
surface station to orbiter\footnote{Orbiters can communicate at approximately
$0.208$-$\SI{0.521}{\mega\bit\per\second}$ with a surface station for $8$ minutes per
sol, or Martian day~\citep{jplnasacomms}.},
and orbiter to Earth\footnote{To transmit from the orbiter to Earth, the communication rate depends on which
orbiter is above the lander to relay the data to Earth. The simulation study
in~\cref{sssec:constrained_bandwidth} assumes the lowest data rate from the Mars
Odyssey orbiter, which ranges from
$0.128$-$\SI{0.256}{\mega\bit\per\second}$~\citep{jplnasacomms}.}.
The bottleneck in communication is between the subterranean robot and
surface station when the robot is transmitting at depths between
\SI{20}{}--\SI{25}{\meter} below ground, so the results
in~\cref{sssec:constrained_bandwidth} are presented for the rates
\SI{0.1}{}--\SI{0.25}{\mega\bit\per\second}, which are in line with data
transmission rates at these depths.  Throughout this section
the shorthand OG is used to refer to the occupancy grid
mapping approach~\citep{elfes1989} while OM refers to
OctoMap~\citep{hornung2013octomap}.

\subsection{Perceptual Detail Evaluation\label{sssec:perceptual_detail}}
\begin{figure}[t]
  \centering
  \subfloat[\label{sfig:crevice_rgb}RGB Image]{\includegraphics[height=2.5cm,trim=70 0 100 0,clip]{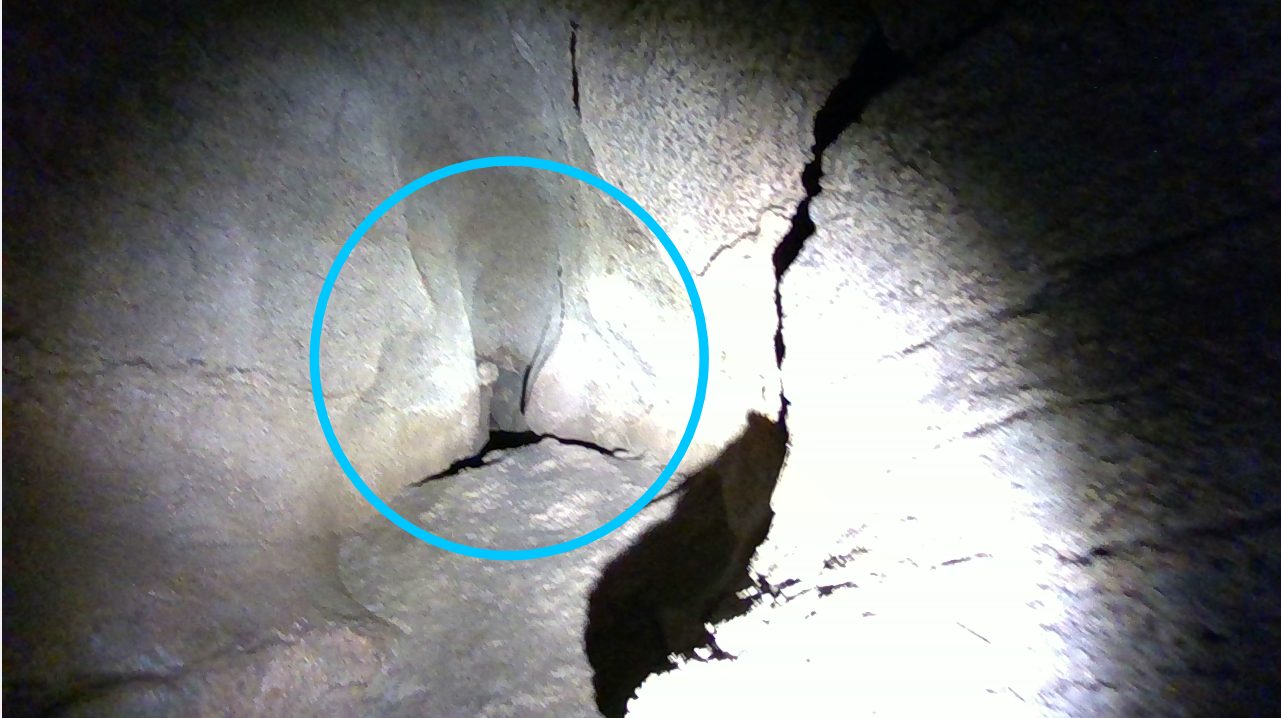}}\hfill
  \subfloat[\label{sfig:crevice_depth}Point Cloud]{\includegraphics[height=2.5cm,trim=0 30 0 30,clip]{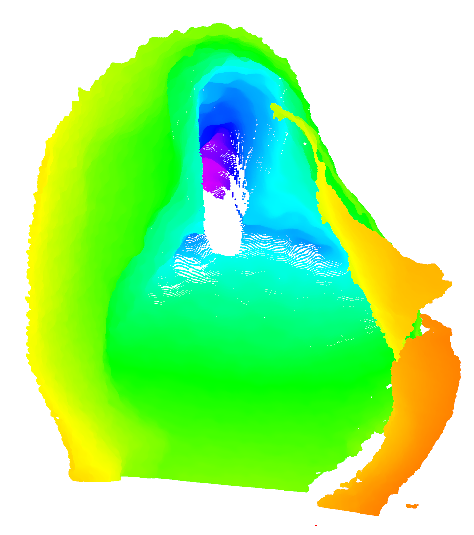}}\hfill
  \subfloat[\label{stab:memory_table}Memory]{\begin{tabular}[b]{c|c|c|c}
    \toprule
      & $\SI{0.025}{\meter}$ & $\SI{0.05}{\meter}$ & $\SI{0.1}{\meter}$ \\
  \hline
        (bytes) & (bytes) & (bytes) & (bytes) \\
  \hline
     \textbf{GMM} & \multicolumn{3}{c}{\textbf{Occupancy Grid (OG)}}\\
    $4028$ & $\SI{1.3e6}{}$  & $\SI{1.8e5}{}$ & $\SI{2.7e4}{}$  \\
  \hline
    \textbf{GMM} & \multicolumn{3}{c}{\textbf{OctoMap (OM)}}\\
    $4028$ & $\SI{2.2e5}{}$ & $\SI{5.8e4}{}$ & $\SI{1.4e4}{}$ \\
    \bottomrule
 \end{tabular}
}\\
  \subfloat[\label{sfig:crevice_gmm} Resampled GMM]{\includegraphics[height=2.75cm,trim=0 50 0 25,clip]{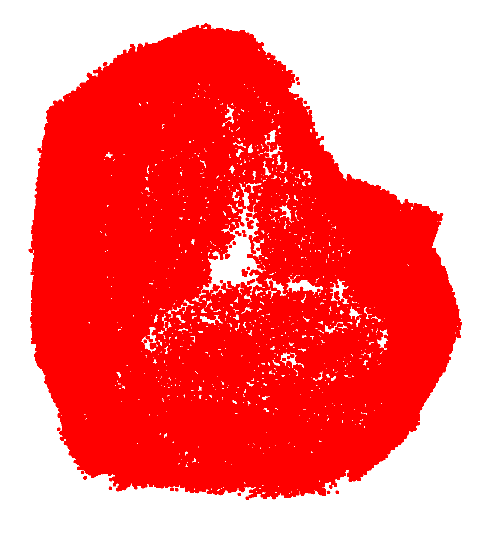}}\hfill
  \subfloat[\label{sfig:crevice_octomap_0.025m} OM ($\SI{0.025}{\meter}$)]{\includegraphics[height=2.75cm,trim=0 0 0 0,clip]{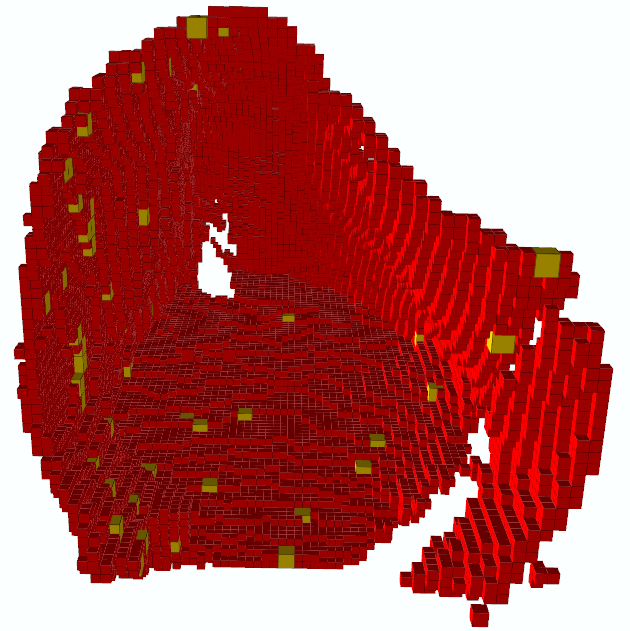}}\hfill
  \subfloat[\label{sfig:crevice_octomap_0.05m} OM ($\SI{0.05}{\meter}$)]{\includegraphics[height=2.75cm,trim=0 0 0 0,clip]{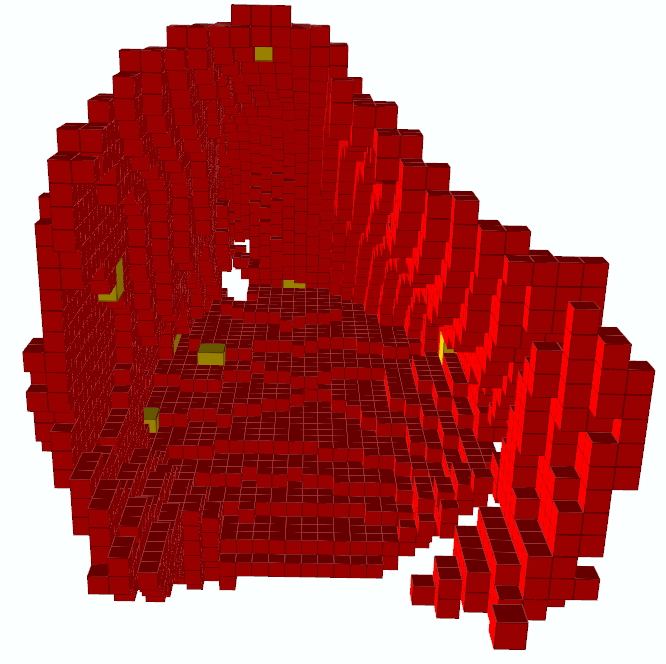}}\hfill
  \subfloat[\label{sfig:crevice_octomap_0.1m} OM ($\SI{0.1}{\meter}$)]{\includegraphics[height=2.75cm,trim=0 0 0 0,clip]{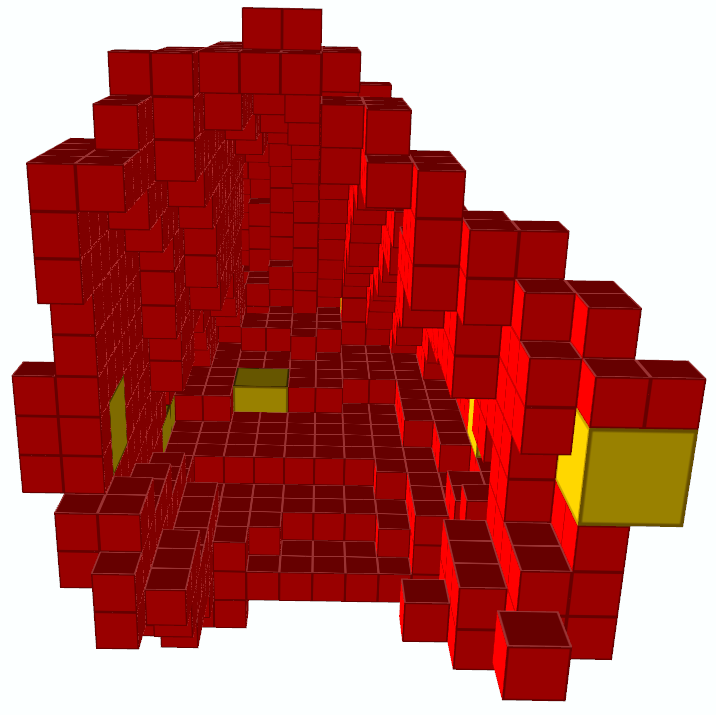}}\\
  \caption{\label{fig:fidelity-comparison}
    Fidelity and memory usage evaluation of several mapping
approaches.  \protect\subref{sfig:crevice_rgb} and
\protect\subref{sfig:crevice_depth} illustrate data from a
representative environment the robot may encounter in the cave. A
potential passage is circled in
cyan.
\protect\subref{stab:memory_table} highlights
significant reduction in memory usage required by the GMM approach as
compared to the OG and OM approaches.~\protect\subref{sfig:crevice_gmm} Resampled points from the GMM
are shown in red.
\protect\subref{sfig:crevice_octomap_0.025m}--\protect\subref{sfig:crevice_octomap_0.1m}
illustrate the OctoMap representation with leaf sizes varying from
\SI{0.025}{\meter} to \SI{0.1}{\meter}. Leaf voxels are shown in
red and larger voxels in yellow.
  }
\end{figure}

The first evaluation compares the perceptual fidelity of different
environment representations in the context of memory usage. An RGB
image and point cloud of a crevice in the cave are shown in
~\cref{sfig:crevice_rgb,sfig:crevice_depth} respectively.
It is not clear from the image and depth information if the
passage continues or there is a lack of data due to insufficient
accuracy in the sensor observation. In either case, additional views
are required to determine the exact nature of the passage.
\Cref{stab:memory_table} demonstrates that as the resolution of the OG and
OM approaches increases, the memory demands also substantially
increase. By comparison, the GMM approach requires substantially less
memory.  When using the GMM approach, the resulting resampled point
cloud is shown in~\cref{sfig:crevice_gmm}, where a hole in the data is
visible. This approach is compared to OM
with varying leaf sizes in~\cref{sfig:crevice_octomap_0.025m,sfig:crevice_octomap_0.05m,sfig:crevice_octomap_0.1m}.

To obtain these results, a GMM was learned consisting of 100 components. Each
component requires \SI{10}{} floating point numbers which includes six
floating point numbers to represent the symmetric covariance, three floating
point numbers for the mean, and one floating point number to represent the
mixing weight. Additional memory was used to represent the pose via six floating
point numbers (three each for translation and rotation) where each floating
point number is assumed to be four bytes. A 32-bit unsigned integer (four bytes)
is also used to represent the support size of the GMM. In the OG case,
one floating point number is used to store the logodds value and one
unsigned integer (four bytes) is used to represent the index for each voxel
in the change set. The total change set of $N$ voxels is transmitted along with
meta-data to reconstruct the grid. The meta-data consists of three unsigned
integers to represent the dimensions of the grid in width, height, and length as
well as three floating point numbers to represent the origin for a total of
\SI{24}{} bytes. The total data required to represent the sensor observation
with an OG is $8N + 24$ bytes. For OM, the full
probabilistic model is serialized and stored to disk. The size of the file is
reported in the table. The motivation for retaining the logodds
values in the OG and OM representations is to enable information-theoretic
planning.
\begin{figure}[t]
  \centering
  \subfloat[\label{sfig:env}Robots (circled) deployed in a cave. Communication
  router shown via dotted line.]{\includegraphics[width=\linewidth,trim=0 0 0 0,clip]{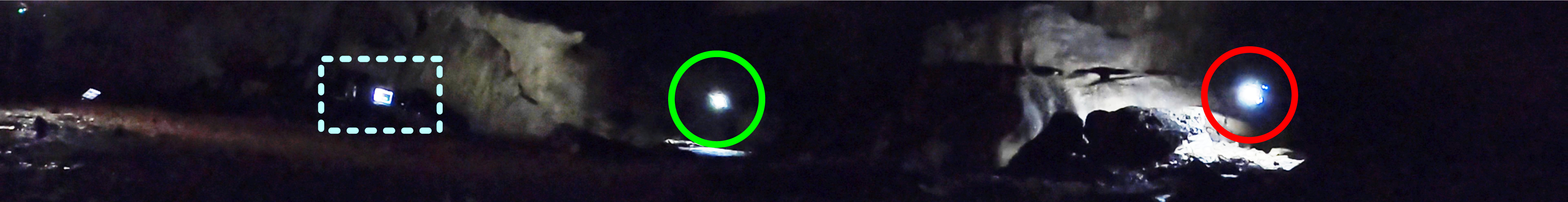}}\\
  \subfloat[\label{sfig:gmm_cave}Combined GMM map]{\includegraphics[width=0.33\linewidth,trim=0 0 0 0,clip]{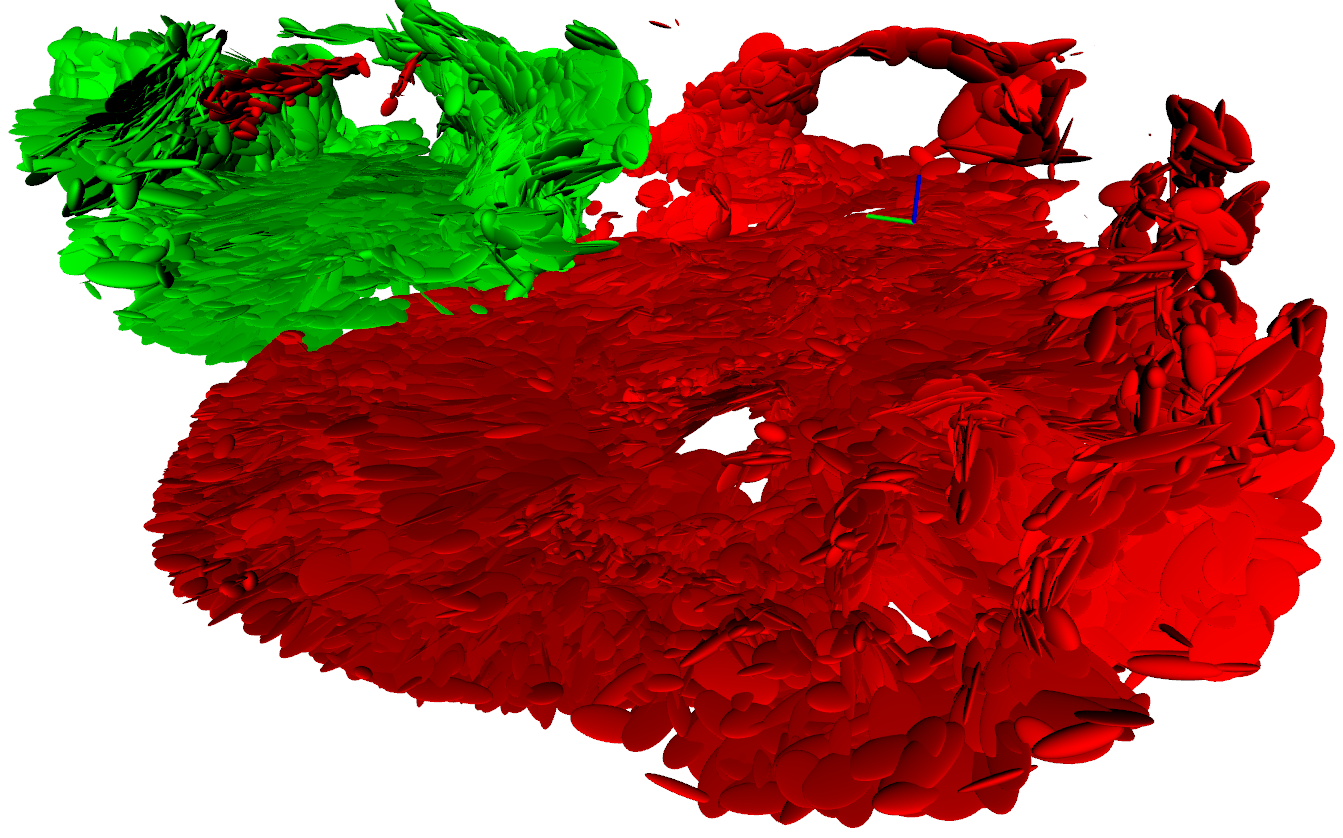}}%
  \subfloat[\label{sfig:speeds_cave}Speed bounds
  shown by dashed lines.]{
\begin{tikzpicture}

\begin{axis}[
width=0.275\textwidth,
colorbar,
colorbar style={at={(1.02,1.0)}, ytick={0,5,10},yticklabels={0\%,5\%,10\%},ylabel={},
  width=0.1*\pgfkeysvalueof{/pgfplots/parent axis width}},
colormap/viridis,
label style={font=\scriptsize},
point meta max=10.0985804618982,
point meta min=0,
tick align=outside,
title=\textbf{\red{R1}},
title style={yshift=-1.5ex},
tick pos=left,
x grid style={lightgray!92.0261437908!black},
xlabel={Speeds (m/s)},
xmin=0.00134342832974514, xmax=2.37057074751781,
xtick style={color=black},
y grid style={lightgray!92.0261437908!black},
ylabel={Yaw Rate (rad/s)},
ymin=4.4651522369734e-07, ymax=0.666699776539189,
ytick style={color=black},
ytick={0,0.2,0.4,0.6},
yticklabels={0.0,0.2,0.4,0.6}
]
\addplot graphics [includegraphics cmd=\pgfimage,xmin=0.00134342832974514, xmax=2.37057074751781, ymin=4.4651522369734e-07, ymax=0.666699776539189] {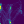};
\path [draw=orange, thick, dash pattern=on 5.55pt off 2.4pt]
(axis cs:1.031,4.4651522369734e-07)
--(axis cs:1.031,0.666699776539189);

\path [draw=red, thick, dash pattern=on 5.55pt off 2.4pt]
(axis cs:2.015,4.4651522369734e-07)
--(axis cs:2.015,0.666699776539189);

\end{axis}

\end{tikzpicture}
\begin{tikzpicture}

\begin{axis}[
width=0.275\textwidth,
colorbar,
colorbar style={at={(1.02,1.0)}, ytick={0,15,30},yticklabels={0\%,15\%,30\%},ylabel={},
  width=0.1*\pgfkeysvalueof{/pgfplots/parent axis width}},
colormap/viridis,
point meta max=36.0825270968406,
point meta min=0,
label style={font=\scriptsize},
tick align=outside,
tick pos=left,
title=\textbf{\green{R2}},
title style={yshift=-1.5ex},
x grid style={lightgray!92.0261437908!black},
xlabel={Speeds (m/s)},
xmin=0.00223458350136284, xmax=2.30765476159221,
xtick style={color=black},
y grid style={lightgray!92.0261437908!black},
ylabel={Yaw Rate (rad/s)},
ymin=3.59799600085694e-05, ymax=0.615516591145312,
ytick style={color=black},
ytick={0,0.2,0.4,0.6},
yticklabels={0.0,0.2,0.4,0.6}
]
\addplot graphics [includegraphics cmd=\pgfimage,xmin=0.00223458350136284, xmax=2.30765476159221, ymin=3.59799600085694e-05, ymax=0.615516591145312] {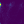};
\path [draw=orange, thick, dash pattern=on 5.55pt off 2.4pt]
(axis cs:1.031,3.59799600085694e-05)
--(axis cs:1.031,0.615516591145312);

\path [draw=red, thick, dash pattern=on 5.55pt off 2.4pt]
(axis cs:2.015,3.59799600085694e-05)
--(axis cs:2.015,0.615516591145312);

\end{axis}

\end{tikzpicture}}\\
  \subfloat[\label{stab:rapps_comm}Communication-Efficiency Comparison]{\input{figs/rapps_table.tex}}
  \caption{\label{fig:multi-robot-exploration} Rapid and communication
    efficient exploration of a cave with a team of two aerial
    robots.~\protect\subref{sfig:env} illustrates the environment with the
    two robots (\textbf{\red{R1}} and \textbf{\green{R2}}) and the WiFi
    router used for communication.~\protect\subref{sfig:gmm_cave}
    illustrates the final GMM maps generated on the
    base-station.~\protect\subref{sfig:speeds_cave}
    shows the percentage density plots
    for linear speeds and yaw rates as measured by the visual-inertial
    navigation system during flight.~\protect\subref{stab:rapps_comm}
    highlights that the GMM approach requires significantly less memory to
    represent the combined map as compared to state-of-the-art
    approaches. In the context of transmitting this data using a channel
    with capacity $\SI{0.25}{\mega\bit\per\second}$, it would take
    significantly less time for the GMM approach as compared to the other
    approaches. A video of the flight can be accessed here: \texttt{https://youtu.be/osko8EKKZUM}.}
\end{figure}

The advantage of the GMM approach is that the probability of occupancy can be
reconstructed at an arbitrary voxel
resolution~\citep{tabibRSS2019,omeadhra2018variable}, which significantly
reduces the memory requirements as compared to the OG and OM approaches. The OG
and OM approaches must retain the probability of occupancy to enable
information-theoretic exploration~\citep{charrow2015icra,zhang2020fsmi}.

\subsection{Hardware Experiments\label{sssec:exploration}}
The second evaluation consists of hardware experiments for two aerial systems
exploring the cave. The experiment demonstrates (1) each robot generates
informative plans with linear speeds up to $\SI{2.37}{\meter\per\second}$ and
yaw rates up to $\SI{0.6}{\radian\per\second}$ while maintaining safety and (2)
the communication required to transmit the map from robots to a base station is
substantially less as compared to the OG and OM approaches. For the purposes of
this experiment, the robots are deployed in disjoint bounding boxes and the
coordination between robots is not studied. What follows is a description of the
experimental setup (including the implementation details) and results.

Each robot in the multi-robot system employs the navigation and
control technique outlined in prior work~\citep{Tabib2020TRO}. The
robots communicate with other computers on the network via WiFi and
use the User Datagram Protocol (UDP) to transfer packets over the
network. Before the start of each experiment,
the SE(3) transform between the takeoff positions of the robots is
measured manually using the navigation approach. The relative
initial transform is used by the distributed mapping subsystem to
align the GMM map fragments in the frames of other robots to the
current robot's local frame.

The maximum speed\footnote{The speed limits and the
operational volumes were chosen based on the cave passage dimensions. The
authors worked with cave management to select a test site that contained neither
actively growing speleothems or bats. Possible effects of imperfect trajectory
tracking and state estimation were also taken into account.} of the robots in the xy-plane is $\SI{2.0}{\meter\per\second}$, the maximum
speed towards unknown space is $\SI{1.0}{\meter\per\second}$, the maximum
z-direction speed is $\SI{0.25}{\meter\per\second}$, and the maximum yaw rate is
constrained to $\SI{0.5}{\radian\per\second}$.
One of the metrics used to assess the planning performance is
quantifying the maximum speed and yaw rate achieved by the robot
while ensuring collision free operation.
Both linear and yawing motions are exploratory actions for an aerial
robot equipped with a limited field of view depth sensor~\citep{goel2019fsr,Tabib2020TRO}.
The data transmitted from the robots to
the base station is used to quantify the success of the mapping approach. The GMM results
of~\cref{fig:multi-robot-exploration} are generated in flight during an actual
trial in the cave. To enable a fair comparison, the depth images collected from
the GMM exploration trial in the cave are post-processed using the OG and OM
approaches. This ensures that variation in the other subsystems does not unduly
affect the results. An analysis to quantify the memory required for each
approach similar to~\cref{sssec:perceptual_detail} is presented. The OG and OM
results are generated by updating the map using the depth information for the
current image and publishing the change set. For the OM approach, the change
set is serialized to file as the full probabilistic model to enable the base
station and other robot to exactly recreate the map for information-theoretic
exploration.

The two deployed robots are denoted by \textbf{\red{R1}} and
\textbf{\green{R2}} in \cref{fig:multi-robot-exploration}. The robots
achieve high exploration rates by selecting actions that enable safe operation at
linear speeds up to $\SI{2.37}{\meter\per\second}$ and yaw rates up to
$\SI{0.6}{\radian\per\second}$, which are
of the same order as state-of-the-art fast exploration
works\footnote{The attained speeds exceed the limits
slightly due to imperfect trajectory tracking and state estimation.}~\citep{goel2019fsr,dharmadhikari2020motion,Dai2020}.
Moreover, note that since \textbf{\red{R1}} operates in a relatively open space
compared to \textbf{\green{R2}}, a larger percentage of high speed actions are
selected (\cref{sfig:speeds_cave}). In contrast, the planner selects the
yawing motion and slow linear actions towards frontiers more often for
\textbf{\green{R2}} to allow for safe operation in a constrained
space (\cref{sfig:speeds_cave}). Both of these behaviors in the multi-robot
system arise automatically due to the choice of the action representation for
single-robot planning in~\citep{goel2019fsr}. These behaviors show that the same
action representation can be used on every robot in the team without any change
in parameters and still allow for intelligent speed adaptation for rapid and
safe exploration.

The combined map from \textbf{\red{R1}} and \textbf{\green{R2}} requires
significantly less time to transmit under the bandwidth constraint when measuring at various
points during exploration (\cref{stab:rapps_comm}). An implication of this in the
context of the concept of operations is that at
$100\%$ exploration completion it will take about $\SI{104.40}{}$
seconds to transmit the GMM map, $\SI{12.30}{}$ hours to transmit the
$\SI{0.025}{\meter}$ resolution OG map, and $\SI{1.25}{}$ days to
transmit the $\SI{0.025}{\meter}$ resolution OM map to Earth. It is
important to note why the OM approach requires more memory than the OG
approach for this
result while it required less memory than the OG approach in~\cref{stab:memory_table}. The change set must be
encoded as an OctoMap before serializing to file. The approach
presented by~\citet{hornung2013octomap} requires that the spatial
relationships between nodes be implicitly stored in the encoding.
This means that the serialized stream does not contain any 3D
coordinates and additional data must be stored to preserve the
structure of the octree. This is in contrast to the OG approach
that stores a logodds value and index from which 3D coordinates can be recovered.
Therefore, for small change sets, the OM approach has much higher
overhead than the OG approach.

\subsection{Effects of Constrained Communication\label{sssec:constrained_bandwidth}}
\begin{figure}[t]
  \centering
  \subfloat[\label{sfig:sim_comms_1}No limit]{\includegraphics[width=0.33\linewidth,trim=0 0 0 0,clip]{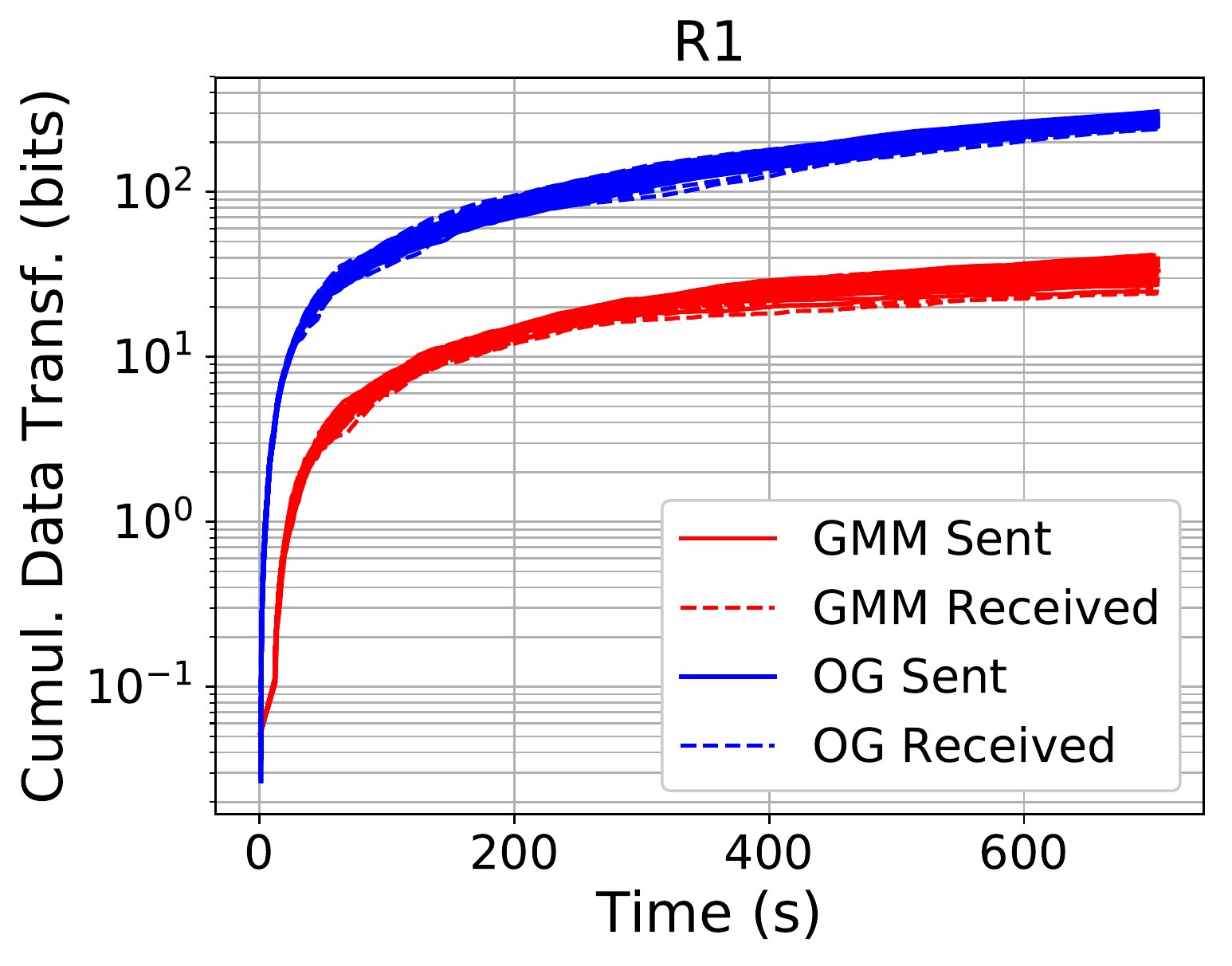}}
  \subfloat[\label{sfig:sim_comms_2}$\SI{0.25}{\mega\bit\per\second}$]{\includegraphics[width=0.33\linewidth,trim=0 0 0 0,clip]{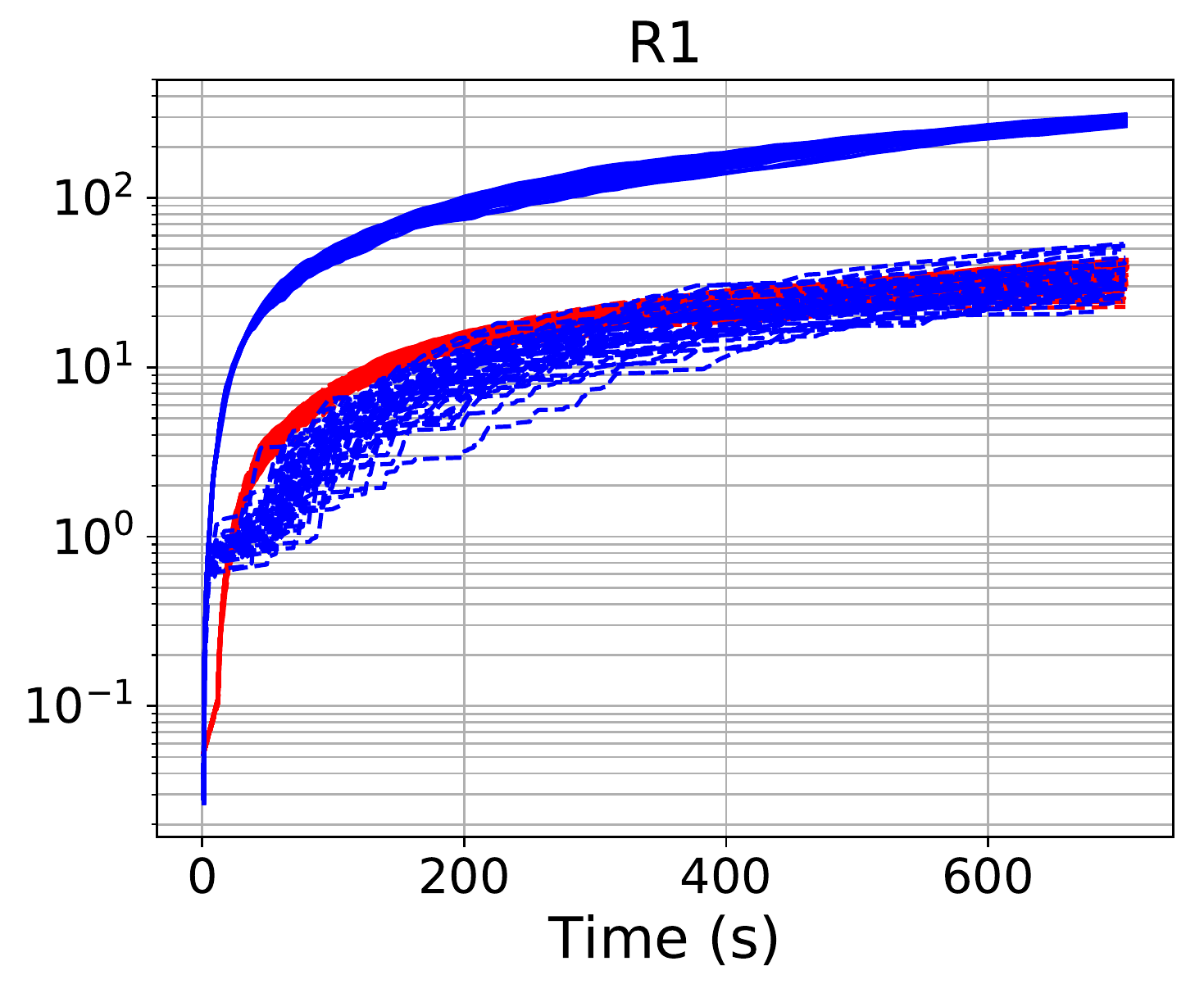}}
  \subfloat[\label{sfig:sim_comms_3}$\SI{0.1}{\mega\bit\per\second}$]{\includegraphics[width=0.33\linewidth,trim=0 0 0 0,clip]{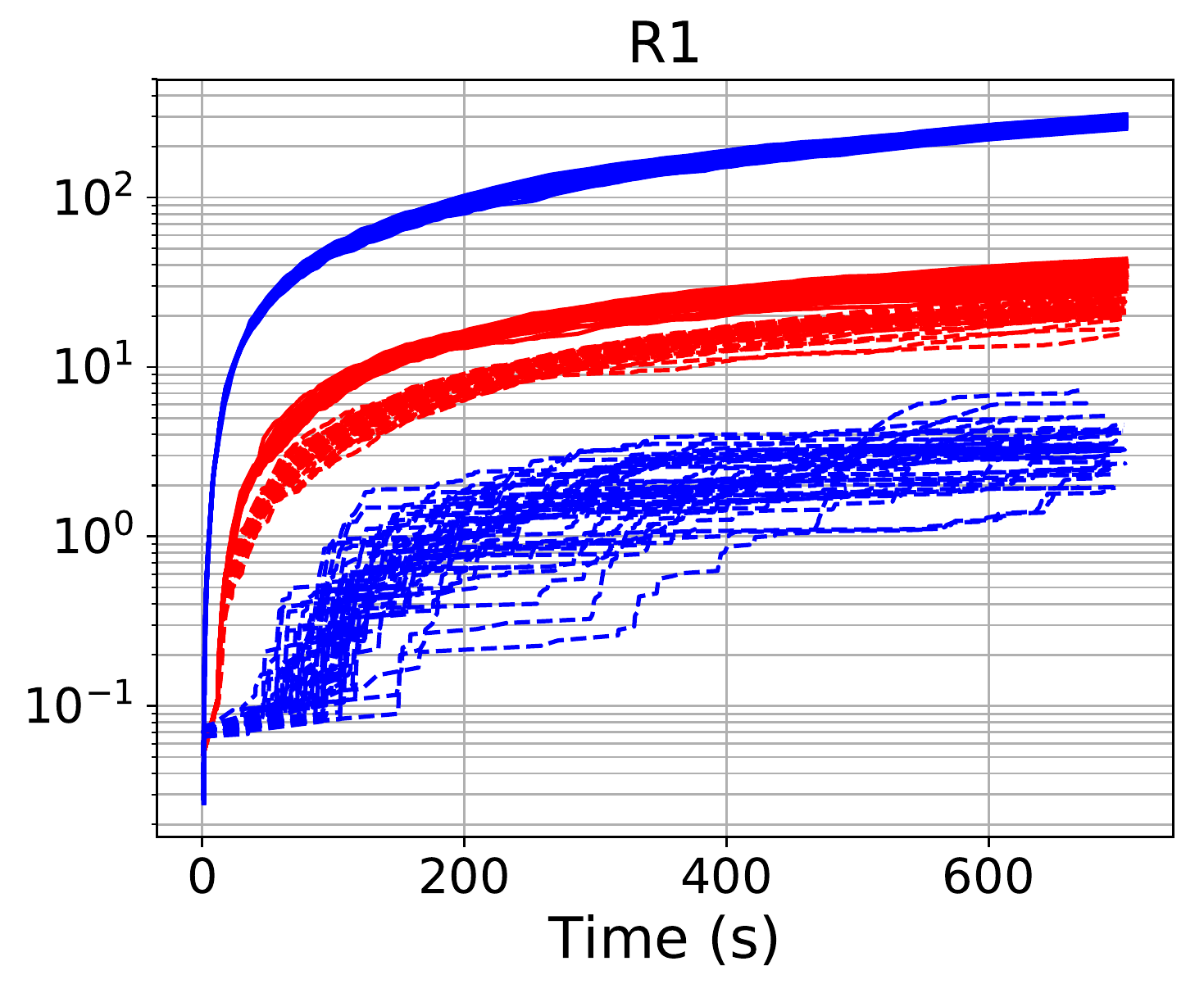}}\\
  \subfloat[\label{stab:sim_table}Exploration completion times]{\input{figs/sim_table.tex}}
  \caption{\label{fig:constrained_bandwidth} Variation of exploration
    performance with inter-robot communication
    limits.~\protect\subref{sfig:sim_comms_1},
    \protect\subref{sfig:sim_comms_2}, and \protect\subref{sfig:sim_comms_3}
    plot the cumulative map data sent and received for the GMM and OG approaches under
    different data rate constraints (the plots are shown for R1 only for brevity).
    The received data is impacted significantly for the OG approach at $\SI{0.25}{\mega\bit\per\second}$
    while both approaches are affected at $\SI{0.1}{\mega\bit\per\second}$. Note that in all
    experiments the planning and coordination methodology is kept the same for a
    fair comparison.~\protect\subref{stab:sim_table} compares
    the time to achieve a certain percentage of environment coverage. We observe that at
    the $\SI{0.25}{\mega\bit\per\second}$ constraint, the GMM approach improves the performance of the team
    by up to $23.84\%$.}
\end{figure}

For this study the assumption on the robots operating in disjoint spaces is
relaxed and a priority-based inter-robot collision checker is implemented for
shared space operation. The simulation consists of a two-robot team that
explores the cave environment. Two approaches are tested: GMM and OG. The
OM approach is not compared for this experiment because to the best of our
knowledge there is no existing open-source implementation of the Shannon mutual
information used for planning by~\citet{zhang2020fsmi}. Further, this enables us
to retain the same planning subsystem for a fair comparison of the GMM and OG approaches.
The communication rate is varied among
\SI{0.1}{\mega\bit\per\second}, \SI{0.25}{\mega\bit\per\second}, and
unconstrained. Each configuration is tested in $40$ experiments with a
$\SI{700}{\second}$ duration. The duration of the exploration is chosen based on the top
speed of the robots and the spatial dimensions of the environment. The
exploration software is run on separate computers in a distributed fashion over
a wired connection.
The simulations are run on two desktop computers running Ubuntu 18.04 with
Intel i7-6700K CPUs. One computer has \SI{32}{\giga\byte} RAM and the other
has \SI{16}{\giga\byte} of RAM.
For the wired connection, the data rate is limited via the network traffic
control tool in Linux that uses the Token Bucket Filter (TBF) to maintain the
specified rate value~\citep{hubert2002linux}. \Cref{fig:constrained_bandwidth}
illustrates the results from the simulation study. As the communication
bandwidth is reduced from no limit in~\cref{sfig:sim_comms_1} to
\SI{0.25}{\mega\bit\per\second} the OG approach begins to drop packets and the
exploration performance of the multi-robot approach decreases as compared to the
GMM approach (see~\cref{stab:sim_table}). At this rate, the GMM approach
achieves 85\% environment coverage in less than 80\% of the time that it takes
the OG approach. However, as the communication rate decreases further to
\SI{0.1}{\mega\bit\per\second} the GMM approach also suffers though it is able
to outperform the OG approach.



\section{Conclusion and Future Work}\label{sec:conclusion}
This work leveraged the compactness of Gaussian mixture models for
high-fidelity perceptual modeling to increase the rate of multi-robot
exploration in reduced bandwidth scenarios such as autonomy in
caves. The mapping approach enables retention of environment details
while remaining amenable to low-bandwidth transmission. The advantage
of this mapping strategy is that it enables a substantial increase in
exploration rate of the multi-robot team as compared to
state-of-the-art mapping techniques even as the communication
bandwidth of the connection between robots decreases.  Future work will
improve perceptual detail in the environment and develop hierarchical
strategies that adapt the fidelity of the model based on the sensor
data. Multi-modal mapping (for example, thermal, RGB, etc.)  may also
be beneficial in these scenarios. Finally, coordination strategies can
be developed to enable robots to share communication-efficient
policies and improve the rate of exploration.

\bibliographystyle{./spbasic}
{
  \bibliography{content/refs}
}

\end{document}